\newif\ifsinglecolumn
\singlecolumntrue

\ifsinglecolumn
    \documentclass[11pt,journal,draftcls,a4paper,onecolumn]{IEEEtran}
\else
    \documentclass[journal,final,letterpaper,twoside,twocolumn]{IEEEtran}
\fi

\usepackage[utf8]{inputenc}
\usepackage[T1]{fontenc}
\usepackage{amsmath}
\usepackage{amssymb}
\usepackage{mathrsfs}
\usepackage{color}
\usepackage{cite}
\usepackage{algorithm,algpseudocode}
\usepackage{graphicx}
\usepackage[labelformat=simple]{subcaption}

\usepackage{multirow}

\algnewcommand\algorithmicinput{\textbf{Input:}}
\algnewcommand\INPUT{\item[\algorithmicinput]}
\algnewcommand\algorithmicoutput{\textbf{Output:}}
\algnewcommand\OUTPUT{\item[\algorithmicoutput]}

\DeclareMathOperator*{\argmin}{arg\,min}

\title{Robust fusion of multi-band images with different spatial and spectral resolutions for change detection}

\author{Vinicius Ferraris, Nicolas Dobigeon, \IEEEmembership{Senior Member, IEEE}, \\Qi Wei, \IEEEmembership{Member, IEEE}, and Marie Chabert
\thanks{Part of this work has been submitted to the IEEE Int. Conf. Acoust., Speech and Signal Process. (ICASSP), 2017 \cite{Ferraris2016icassp}.}
\thanks{Part of this work has been supported by Coordenação de Aperfeiçoamento de Ensino Superior (CAPES), Brazil, and EU FP7 through the ERANETMED JC-WATER Program, MapInvPlnt Project ANR-15-NMED-0002-02.}
\thanks{V. Ferraris, N. Dobigeon and M. Chabert are with University of Toulouse, IRIT/INP-ENSEEIHT, France  (email: \{vinicius.ferraris, nicolas.dobigeon, marie.chabert\}@enseeiht.fr).}
\thanks{Q. Wei is with Department of Engineering, University of Cambridge, CB2 1PZ, Cambridge, UK (email: qw245@cam.ac.uk).}
}

\begin{document}
\maketitle

\begin{abstract}

Archetypal scenarios for change detection generally consider two images acquired through sensors of the same modality. However, in some specific cases such as emergency situations, the only images available may be those acquired through different kinds of sensors. More precisely, this paper addresses the problem of detecting changes between two multi-band optical images characterized by different spatial and spectral resolutions. This sensor dissimilarity introduces additional issues in the context of operational change detection. To alleviate these issues, classical change detection methods are applied after independent preprocessing steps (e.g., resampling) used to get the same spatial and spectral resolutions for the pair of observed images. Nevertheless, these preprocessing steps tend to throw away relevant information. Conversely, in this paper, we propose a method that more effectively uses the available information by modeling the two observed images as spatial and spectral versions of two (unobserved) latent images characterized by the same high spatial and high spectral resolutions. As they cover the same scene, these latent images are expected to be globally similar except for possible changes in sparse spatial locations. Thus, the change detection task is envisioned through a robust multi-band image fusion method which enforces the differences between the estimated latent images to be spatially sparse. This robust fusion problem is formulated as an inverse problem which is iteratively solved using an efficient block-coordinate descent algorithm. The proposed method is applied to real panchormatic/multispectral and hyperspectral images with simulated realistic changes. A comparison with state-of-the-art change detection methods evidences the accuracy of the proposed strategy.
\end{abstract}



\renewcommand{\thesection}{\Roman{section}}

\section{Introduction}

\PARstart{R}{emote} sensing is a reliable technique for Earth surface monitoring and observation \cite{elachiintroduction2006,campbellintroduction2011}. One of the most important applications using remotely sensed data is the so-called change detection (CD) problem. CD has many definitions and it is generally considered as the ability of analyzing two or more multi-date (i.e., acquired at different time instants) and possibly multi-source (i.e., acquired by different sensors) images of the same scene to detect areas where potential changes have occurred \cite{singhreview1989,bovolotime2015}. Because of the increasing number of satellites and of new policies for data distribution, more multi-temporal data becomes available. While it increases the amount of information on the present scene, it highlights some additional issues when designing operational change detection techniques.

Each remotely sensed observation image is intimately connected to the acquisition modality providing a particular excerpt of the observed scene according to the sensor specifications. For instance, optical images are generally well suited to map horizontal structures, e.g., land-cover type at large scales  \cite{dalla_murachallenges2015}. More particularly, remote sensing images acquired by multi-band optical sensors can be classified according to their spectral and spatial resolutions. The spectral resolution is related to the capability in sensing the electromagnetic spectrum. This term can also refer to the number of spectral bands \cite{landgrebehyperspectral2002,campbellintroduction2011}, which generally leads to a commonly adopted classification of these images: \textit{panchromatic} (PAN) images, characterized by a low spectral resolution, \textit{multispectral} (MS) and \textit{hyperspectral} (HS) images which sense part of the spectrum with higher precision. Alternatively, multi-band optical images can be classified with respect to (w.r.t.) their spatial resolution \cite{dalla_murachallenges2015,campbellintroduction2011}. The concept of spatial resolution should be understand as the capability of representing the smallest object that can be resolved up to a specific pixel size. Images having small resolution size and finer details are generally identified as of \textit{(high resolution} (HR) in contrast to \textit{low resolution} (LR) images where only coarse features are observable. Because of the physical limitations of optical passive sensors, multi-band optical images suffer from a trade-off between spectral and spatial resolution \cite{pricespectral1997,elachiintroduction2006}. To ensure that any sensor has sufficient amount of energy to guarantee a proper acquisition (in terms of, e.g., signal-to-noise ratio), one of the resolutions must be decreased allowing the other to be increased. For this reason, PAN images are generally characterized by higher spatial resolution and lower spectral resolution than MS or a HS images.

Optical images have been the most studied remote sensing modality for CD since the widely admitted additive Gaussian modeling of optical optical sensor noises allows CD techniques to be implemented through a simple operation of image differencing \cite{singhreview1989,bovolotime2015}. Originally designed for single-band images, CD differencing methods have been adapted to handle multi-band images by considering spectral change vectors \cite{bovolotheoretical2007,bovoloframework2012} and transform analysis \cite{nielsenmultivariate1998,nielsenregularized2007}. The possibility of detecting changes by exploiting both spatial and spectral information is one of the greatest advantages of these multi-band images. Nevertheless, images of same modality are not always available. In some specific scenarios, for instance consecutive to natural disasters, the availability of data imposes observation images acquired through different kind of sensors. Such disadvantageous emergency situations yet require fast, flexible and accurate methods able to handle also the incompatibilities introduced by the each sensor modality \cite{ingladasimilarity2002,albergaperformance2007,mercierconditional2007,prendesnew2015}. Most of the CD classical methods do not support differences in resolutions. Generally, each observed image is independently preprocessed in order to get the same resolution and then classical CD techniques are applied. However, independent resampling operations do not take into account the pair of observed images and even throw away important information. Recently, a general CD framework has been proposed in \cite{ferrarisdetecting2016} to deal with multi-band images with different spatial and spectral resolutions based on a $3$-step procedure (fusion, prediction, detection). Instead of independently preprocessing each observed image, this approach consists in recovering a latent (i.e., unobserved) HR-HS image containing changed and unchanged regions by fusing both observed images. Then, it predicts pseudo-observed images by artificially degrading the estimated HR-HS latent image using the same forward models underlying the actually observed images. As the pairs of predicted and observed observations have the same spatial and spectral resolutions, any classical multi-band CD method can be finally applied to build a change map. Albeit significantly improving detection performance when compared to crude methods relying on independent preprocessing, the $3$-step sequential formulation appears to be non-optimal for the following twofold reasons: i) any inaccuracies in the fusion step are propagated throughout the subsequent degradation and detection steps, ii) relevant information regarding the change may be lost during the prediction steps, since it consists in spatially or spectrally degrading the latent images to estimate the pseudo-observed images. Thus, significant improvements in terms of change detection performance may be expected provided one is able to overcome both limitations.

In this paper, capitalizing on the general framework developed in \cite{ferrarisdetecting2016}, we show that the CD task can be formulated as a particular instance of the multi-band image fusion problem. However, contrary to the $3$-step procedure in \cite{ferrarisdetecting2016}, the proposed approach jointly estimates a couple of distinct HR-HS latent images corresponding to the two acquisition times as well as the change image. Since the two HR-HS latent images are supposed to represent the same scene, they are expected to share a high level of similarity or, equivalently, to differ only in a few spatial locations. Thus, akin to numerous robust factorizing models such as robust principal component analysis \cite{candesrobust2011} and robust nonnegative matrix factorization \cite{Fevotte2015}, the two observed images are jointly approximated by a standard linear decomposition model complemented with an HR-HS outlier term corresponding to the change image. This so-called CD-driven \emph{robust} fusion of multi-band images is formulated as an inverse problem where, in particular, the outlier term is characterized by a spatial sparsity-inducing regularization. The resulting objective function is solved through the use of a block coordinate descent (BCD) algorithm, which iteratively optimizes w.r.t. one latent image and the change image. Remarkably, optimizing w.r.t. the latent image boils down to a classical multi-band image fusion step and can be efficiently conducted following the algorithmic solutions proposed in \cite{weifast2015-2}. The CD map can be finally generated from the recovered HR-HS change image.

The paper is organized as follows. Section \ref{sec:ps} formulates the change detection problem for multi-band optical image. Section \ref{sec:CD} presents the solution for the formulated problem based on robust fusion. The simulation strategy as well as the results and considerations are present in Section \ref{sec:experiments}.

\section{From change detection to robust fusion}
\label{sec:ps}

\subsection{Generic forward model}
\label{subsec:forward}	
Let us consider the image formation process as a sequence of transformations, denoted $T\left[\cdot\right]$, of the original scene into an output image. The output image of a particular sensor is referred to as the observed image and denoted $\mathbf{Y} \in \mathbb{R}^{n_{\lambda}\times m}$ where $m$ and $n_{\lambda}$ are the numbers of pixels and spectral bands in the observed image, respectively. It provides a limited version of the original scene with characteristics imposed by the image signal processor (ISP) characterizing the sensor. The original scene can be conveniently represented by an (unknown) latent image of higher spatial and spectral resolutions, $\mathbf{X} \in \mathbb{R}^{m_{\lambda}\times n}$, where $n \geq m$ and $m_{\lambda}\geq n_{\lambda}$ are the numbers of pixels and spectral bands, respectively, related to the observed image following
	\begin{equation}
	\mathbf{Y} = T\left[\mathbf{X}\right].
	\label{eq:transformation}
	\end{equation}
The intrinsic sequence of transformations of the sensor over the latent image $\mathbf{X}$ can be typically classified as spectral or spatial degradations. On one hand, spatial degradations are related to the spatial characteristics of the sensor such as sampling scheme and optical transfer function. On the other hand, spectral degradations refer to the wavelength sensitivity and the spectral sampling. There are many ways to represent the degradation process. In this paper, is is  considered as a sequence of linear operations leading to the following generic forward model \cite{weibayesian2015-2,Yokoya2012,Simoes2014b}
    \begin{equation}
	\label{eq:model}
		\mathbf{Y} = \mathbf{L}\mathbf{X}\mathbf{R} + \mathbf{N}
	\end{equation}
	where
	\begin{itemize}
		\item $\mathbf{L} \in \mathbb{R}^{n_{\lambda} \times m_{\lambda}}$ is the spectral degradation matrix,
		\item $\mathbf{R} \in \mathbb{R}^{n \times m}$  is the spatial degradation matrix,
		\item $\mathbf{N}$ is the additive term comprising sensor noise and modeling errors.
	\end{itemize}

In \eqref{eq:model}, the left-multiplying matrix $\mathbf{L} \in \mathbb{R}^{n_{\lambda}\times m_{\lambda}}$ degrades the latent image by combination of some spectral bands for each pixel while the right-multiplying matrix $\mathbf{R} \in \mathbb{R}^{n\times m}$ degrades the latent image by linear combination of pixels within the same spectral band. The former degradation corresponds to a spectral resolution reduction with respect to the latent image $\mathbf{X}$ as in \cite{Yokoya2012,Simoes2014b,weifast2015-2}. In practice, this degradation models an intrinsic characteristic of the sensor, namely the spectral response. It can be either learned by cross-calibration or known \textit{a priori} \cite{Simoes2014b,yokoyacross-calibration2013}. Conversely, the spatial degradation matrix $\mathbf{R}$ models the combination of different transformations which are specific of the sensor architecture taking into account external factors including wrap, blurring, translation and decimation \cite{yokoyacross-calibration2013,heideflexisp:2014,weifast2015-2}. In this work, since geometrical transformations such as wrap and translations can be corrected using image co-registration techniques in pre-processing steps, only a spatially invariant blurring and a decimation (i.e., subsampling) will be considered. A space-invariant blur can be modeled by a symmetric convolution kernel associated with a sparse symmetric Toeplitz matrix $\mathbf{B} \in \mathbb{R}^{n\times n}$ which operates a cyclic convolution on the each individual band \cite{weihyperspectral2015}. The decimation operation, denoted by the $n\times m$ matrix $\mathbf{S}$, corresponds to a uniform downsampling operator\footnote{The corresponding operator $\mathbf{S}^{T}$ represents an upsampling transformation by zero-interpolation from $m$ to $n$.} of factor $d = d_{r} \times d_{c}$ with $ m = n/d$ ones on the block diagonal and zeros elsewhere, such that $\mathbf{S}^{T}\mathbf{S} = \mathbf{I}_{m}$ \cite{weifast2015-2}. To summarize, the overall spatial degradation process corresponds to the matrix composition $\mathbf{R} = \mathbf{B}\mathbf{S} \in \mathbb{R}^{n\times m}$.

The noise corrupting multi-band optical images is generally modeled as additive and Gaussian \cite{bovolotime2015,elachiintroduction2006,loncanhyperspectral2015,weifast2015-2}. Thus the noise matrix $\mathbf{N}$ in \eqref{eq:model} is assumed to be distributed according to the following matrix normal distribution\footnote{The probability density function $p(\mathbf{X}|\mathbf{M},\mathbf{\Sigma}_{r},\mathbf{\Sigma}_{r})$ of a matrix normal distribution $\mathcal{M}\mathcal{N}_{r,c}(\mathbf{M},\mathbf{\Sigma}_{r},\mathbf{\Sigma}_{c})$ is given by \cite{guptamatrix1999}
\begin{center} $ p\left(\mathbf{X}|\mathbf{M},\mathbf{\Sigma}_{r},\mathbf{\Sigma}_{r}\right) = \frac{ \exp \left(-\frac{1}{2}tr \left[
\mathbf{\Sigma}_{c}^{-1} \left(\mathbf{X}-\mathbf{M}\right)^{T} \mathbf{\Sigma}_{r}^{-1} \left(\mathbf{X}-\mathbf{M}\right) \right]\right)}{\left(2\pi\right)^{rc/2}\left|\mathbf{\Sigma}_{c}\right|^{r/2}\left|\mathbf{\Sigma}_{r}\right|^{c/2}}$
\end{center}
where $\mathbf{M} \in \mathbb{R}^{r\times c}$ is the mean matrix, $\mathbf{\Sigma}_{r} \in \mathbb{R}^{r\times r}$  is the row covariance matrix and $\mathbf{\Sigma}_{c} \in \mathbb{R}^{c\times c}$ is the column covariance matrix.}
	\begin{equation}
        \label{eq:noise_stats}
			\mathbf{N} \sim \mathcal{M}\mathcal{N}_{n_{\lambda},m}(\mathbf{0}_{n_{\lambda}\times m},\mathbf{\Lambda},\mathbf{\Pi}).
	\end{equation}
The row covariance matrix $\mathbf{\Lambda}$ carries information regarding the between-band spectral correlation. Following \cite{weifast2015-2}, in what follows, this covariance matrix $\mathbf{\Lambda}$ will be assumed to be diagonal, which implies that the noise is independent from one band to the other and characterized by a specific variance in each band. Conversely, the column covariance matrix $\mathbf{\Pi}$ models the noise correlation w.r.t. to the pixel locations. Following a widely admitted hypothesis of the literature, this matrix is assumed to be identity, $\mathbf{\Pi}=\mathbf{I}_{m}$, to reflect the fact the noise is spatially independent. In real applications, both matrices $\mathbf{\Lambda}$ and $\mathbf{\Pi}$ can be estimated by previous calibrations \cite{yokoyacross-calibration2013}.

\subsection{Problem statement}
\label{subsec:ps}

Let us denote $t_j$ and $t_i$ the acquisition times of two co-registered multi-band optical images. It is not assumed any specific information about time ordering, either $t_i<t_j$ or $t_i>t_j$ are possible cases. Hence, without loss of generality, the HR-PAN/MS image acquired at time $t_i$ is assumed to be a low spectral resolution (i.e., PAN or MS) image of high spatial resolution denoted $\mathbf{Y}_{\mathrm{HR}}^{t_i} \in \mathbb{R}^{n_{\lambda}\times n}$. The image acquired at time $t_j$ is a LR-HS image denoted $\mathbf{Y}_{\mathrm{LR}}^{t_j} \in \mathbb{R}^{m_{\lambda}\times m}$. The problem addressed in this paper consists of detecting significant changes between these two images. This is a challenging task mainly due to the spatial and spectral resolution dissimilarity which prevents any use of simple yet efficient differencing operation \cite{singhreview1989,bovolotime2015}. To alleviate this issue, this work proposes to generalize the CD framework introduced in \cite{ferrarisdetecting2016}. More precisely, following the widely admitted forward model described in Section \ref{subsec:forward} and adopting consistent notations, the observed images $\mathbf{Y}_{\mathrm{HR}}^{t_i}$ and $\mathbf{Y}_{\mathrm{LR}}^{t_j}$ can be related to two HR-HS latent images $\mathbf{X}^{t_i}$ and $\mathbf{X}^{t_j}$, respectively, as follows
\begin{subequations}
\label{eq:jointobsmodel}
		\begin{align}
			&\mathbf{Y}_{\mathrm{HR}}^{t_i} = \mathbf{L}\mathbf{X}^{t_i} + \mathbf{N}_{\mathrm{HR}}  \label{eq:jointobsmodelHR}\\
			&\mathbf{Y}_{\mathrm{LR}}^{t_j} = \mathbf{X}^{t_j}\mathbf{BS} + \mathbf{N}_{\mathrm{LR}} \label{eq:jointobsmodelLR}.
		\end{align}
\end{subequations}
Note that \eqref{eq:jointobsmodelHR} and \eqref{eq:jointobsmodelLR} are a specific double instance of \eqref{eq:model}. Indeed, the HR-PAN/MS (resp., LR-HS) image $\mathbf{Y}_{\mathrm{HR}}^{t_i}$ (resp., $\mathbf{Y}_{\mathrm{LR}}^{t_j}$) is assumed to be only a spectrally (resp., spatially) degraded version of the HR-HS latent image $\mathbf{X}^{t_i}$ (resp., $\mathbf{X}^{t_i}$) such that both latent images $\mathbf{X}^{t_i}\in \mathbb{R}^{m_{\lambda}\times n}$ and $\mathbf{X}^{t_j} \in \mathbb{R}^{m_{\lambda}\times n}$ share the same spectral and spatial resolutions which correspond to the highest resolutions of both observed images. Thereby, provided these two latent images can be efficiently inferred, any classical differencing technique can be subsequently implemented on them to detect changes, notably at a high spatial resolution. More specifically, it would consist of evaluating an HR-HS change image denoted $\Delta\mathbf{X}=\left[\Delta\mathbf{x}_1,\ldots,\Delta\mathbf{x}_n\right]$ that would gather information related to any change between the two observed images
\begin{equation}
\label{eq:assumption}
 \Delta\mathbf{X} = \mathbf{X}^{t_i} - \mathbf{X}^{t_j}
\end{equation}
where $\Delta\mathbf{x}_p\in\mathbb{R}^{m_\lambda}$ denotes the spectral change vector in the $p$th pixel ($p=1,\ldots,n$). This spectral change image can be exploited by conducting a pixel-wise change vector analysis (CVA) \cite{johnsonchange1998} which exhibits the polar coordinates (i.e., magnitude and direction) of the spectral change vectors. To spatially locate the changes, a natural approach consists of monitoring the information contained in the magnitude part of this representation \cite{Singh1989, bovolotheoretical2007, bovoloframework2012}, by considering the corresponding HR spectral change energy image
\begin{equation}
\label{eq:spectral_change_energy_image}
  \mathbf{e} =\left[e_1,\ldots,e_n\right]\in \mathbb{R}^{n}
\end{equation}
with
\begin{equation}
  e_p = \left\|\Delta\mathbf{x}_p\right\|_2, \quad p=1,\ldots,n.
\end{equation}
When the CD problem in the $p$th pixel is formulated as the binary hypothesis testing
\begin{equation}
\label{eq:test}
 \left\{
		\begin{array}{rcl}
			\mathcal{H}_{0,p} &:& \text{no change occurs in the $p$th pixel}  \\
			\mathcal{H}_{1,p} &:& \text{a change occurs in the $p$th pixel}
		\end{array}
        \right.
\end{equation}
the pixel-wise statistical test can be written for a given threshold $\tau$ as
\begin{equation}
    \label{eq:decision_rule}
  e_p \overset{\mathcal{H}_{1,p}}{\underset{\mathcal{H}_{0,p}}{\gtrless}} \tau.
\end{equation}
The final binary HR CD map denoted ${\mathbf{d}} = \left[d_1,\ldots,d_n\right] \in \{0,1\}^n$ can be derived as
	\begin{equation}
	\label{eq:CVArule}
 {d}_p = \left\{\begin{array}{lll}
             1 & \mbox{if } e_p \geq \tau & (\mathcal{H}_{1,p})\\
			 0 & \mbox{otherwise}          & (\mathcal{H}_{0,p}).
				\end{array}\right.
\end{equation}
When complementary information needs to be extracted from the change image $\Delta\mathbf{X}$, e.g., to identify different types of changes, the whole polar representation (i.e., both magnitude and direction) can be fully exploited \cite{bovolotheoretical2007, bovoloframework2012}. As a consequence, to solve the multi-band image CD problem, the key issue lies in the joint estimation of the pair of HR-HS latent images $\left\{\mathbf{X}^{t_i},\mathbf{X}^{t_j}\right\}$ from the forward model \eqref{eq:jointobsmodel} or, equivalently, the joint estimation of one of this latent image and the difference image, e.g., $\left\{\mathbf{X}^{t_j},\Delta\mathbf{X}\right\}$. The next paragraph shows that this problem can be formulated as a particular instance of multi-band image fusion.

\subsection{Robust multi-band image fusion}
\label{subsec:robust_fusion}
Linear forward models similar to \eqref{eq:jointobsmodel} have been extensively investigated in the image processing literature for various applications. When a unique LR-HS image $\mathbf{Y}_{\mathrm{LR}}^{t_j}$ has been observed at time $t_j$, recovering the HR-HS latent image $\mathbf{X}^{t_j}$ from the direct model \eqref{eq:jointobsmodelLR} can be cast as a superresolution problem \cite{JYW2010,zhaofast2016}. Besides, when a complementary HR-PAN/MS image $\mathbf{Y}_{\mathrm{HR}}^{t_i}$ of lower spectral resolution (i.e., PAN or MS) has been simultaneously acquired at time $t_i=t_j$ under \eqref{eq:jointobsmodelHR}, the two corresponding latent images are expected to represent exactly the same scene, i.e., $\Delta\mathbf{X}=\boldsymbol{0}$ or, equivalently, $\mathbf{X}^{t_i} = \mathbf{X}^{t_j} = \mathbf{X}$ where the time index can be omitted. In such scenario, estimating the common HR-HS latent image $\mathbf{X}$ from the two observed images $\mathbf{Y}_{\mathrm{HR}}$ and $\mathbf{Y}_{\mathrm{LR}}$ is a multi-band image fusion problem addressed in \cite{Hardie2004,Eismann2005,Zhang2009,Yokoya2012,Simoes2014b,weibayesian2015-2,weihyperspectral2015,weifast2015-2}, also referred to as MS or HS pansharpening in some specific cases \cite{loncanhyperspectral2015}. Whether the problem consists in increasing the resolution of a single image or fusing multiple images of different spatial and spectral resolutions, the underlying objective consists in compensating the energy trade-off of optical sensors to get highly spatially and spectrally resolved images. Those problems are often formulated as an inverse problem, which is generally ill-posed or, at least, ill-conditioned. To overcome this issue, a classical approach consists of penalizing the data fitting terms derived from the linear models \eqref{eq:jointobsmodel} and the noise statistics \eqref{eq:noise_stats} with additional regularizing terms exploiting any prior information on the latent image. Various penalizations have been considered in the literature, including Tikhonov regularizations expressed in the image domain \cite{Ebrahimi2008,weibayesian2015-2} or a in a transformed (e.g., gradient) domain \cite{YWTai_CVPR_2010,SunJ_TIP_2011}, dictionary- or patch-based regularizations \cite{JYW2010,weihyperspectral2015}, total variation (TV) \cite{SR_Aly2005,Simoes2014b} or regularizations based on sparse wavelet representations \cite{JijiCV2004,GEM_Bioucas-Dias2006}.

In this work, we propose to follow a similar route by addressing, in a first step, the CD problem as a linear inverse problem derived from \eqref{eq:jointobsmodel}. However, the CD problem addressed here differs from the computational imaging problems discussed above by the fact that two distinct HR-HS latent images $\mathbf{X}^{t_i}$ and $\mathbf{X}^{t_j}$ need to be inferred, which makes the inverse problem highly ill-posed. However, this particular applicative scenario of CD yields a natural reparametrization where relevant prior knowledge can be conveniently exploited. More precisely, since the two HR-HS latent images are related to the same scene observed at two time instants, they are expected to share a high level of similarity, i.e., the change image $\Delta\mathbf{X}$ is expected to be spatially sparse. Thus, instead of jointly estimating the pair $\left\{\mathbf{X}^{t_i},\mathbf{X}^{t_j}\right\}$ of HR-HS latent images, we take benefit from this crucial information to rewrite the joint observation model \eqref{eq:jointobsmodel} as a function of $\left\{\mathbf{X}^{t_j},\Delta\mathbf{X}\right\}$, i.e.,
\begin{subequations}
\label{eq:jointobsmodel_bis}
		\begin{align}
			&\mathbf{Y}_{\mathrm{HR}}^{t_i} = \mathbf{L}\left(\mathbf{X}^{t_j}+\Delta\mathbf{X}\right) + \mathbf{N}_{\mathrm{HR}} \label{eq:jointobsmodel_bisHR}\\
			&\mathbf{Y}_{\mathrm{LR}}^{t_j} = \mathbf{X}^{t_j}\mathbf{BS} + \mathbf{N}_{\mathrm{LR}} \label{eq:jointobsmodel_bisLR}.
		\end{align}
\end{subequations}
It is worthy to note that this dual observation model parametrized by the new pair $\left\{\mathbf{X}^{t_j},\Delta\mathbf{X}\right\}$ of images to be inferred can be straightforwardly associated with a particular instance of the multi-band image fusion discussed earlier. Indeed, given the HR-HS change image $\Delta\mathbf{X}$ and the HR-PAN/MS image observed at time $t_i$, an HR-PAN/MS \emph{corrected} image denoted ${\mathbf{Y}}_{\mathrm{cHR}}^{t_j}$ that would be acquired by the HR-PAN/MS sensor at time $t_j$ can be defined as
\begin{equation}
	\label{eq:pseudoObs}
		{\mathbf{Y}}_{\mathrm{cHR}}^{t_j} = \mathbf{Y}_{\mathrm{HR}}^{t_i} - \mathbf{L}\Delta\mathbf{X}.
	\end{equation}
In such case, the HR forward model \eqref{eq:jointobsmodel_bisHR} can be easily rewritten, leading to
\begin{subequations}
\label{eq:jointobsmodel_ter}
\begin{align}
			&{\mathbf{Y}}_{\mathrm{cHR}}^{t_i} = \mathbf{L}\mathbf{X}^{t_j} + \mathbf{N}_{\mathrm{HR}}\\
            &\mathbf{Y}_{\mathrm{LR}}^{t_j} = \mathbf{X}^{t_j}\mathbf{BS} + \mathbf{N}_{\mathrm{LR}}.
\end{align}
\end{subequations}
This observation model \eqref{eq:jointobsmodel_ter} defines a standard multi-band image fusion problem for the LR-HS observed image $\mathbf{Y}_{\mathrm{LR}}^{t_j}$ and the corrected HR-PAN/MS image ${\mathbf{Y}}_{\mathrm{cHR}}^{t_j}$. Consequently, since the change image $\Delta\mathbf{X}$ can be considered as an outlier term, akin to those encountered in several robust factorizing models such as robust principal component analysis (RPCA) \cite{candesrobust2011} and robust nonnegative factorization \cite{Fevotte2015} which relies on a similar sparse outlier term, the joint observation model \eqref{eq:jointobsmodel_bis} naturally defines a so-called \emph{robust fusion} scheme whose objective function is detailed in the next paragraph.

\subsection{Robust fusion objective function}
	
Because of the additive nature and the statistical properties of the noise $\mathbf{N}_{\mathrm{HR}}$ and $\mathbf{N}_{\mathrm{LR}}$, both observed images $\mathbf{Y}_{\mathrm{HR}}^{t_i}$ and $\mathbf{Y}_{\mathrm{LR}}^{t_j}$ can be assumed matrix normally distributed
	\begin{equation*}
		\begin{array}{ccl}
			\mathbf{Y}_{\mathrm{HR}}^{t_i}|\mathbf{X}^{t_j},\Delta\mathbf{X} &\sim& \mathcal{M}\mathcal{N}_{n_{\lambda},n}\left(\mathbf{L}\left(\mathbf{X}^{t_j}+\Delta\mathbf{X}\right),\mathbf{\Lambda}_{\mathrm{HR}},\mathbf{I}_{n}\right) \\
			\mathbf{Y}_{\mathrm{LR}}^{t_j}|\mathbf{X}^{t_j} &\sim& \mathcal{M}\mathcal{N}_{m_{\lambda},m}\left(\mathbf{X}^{t_j}\mathbf{BS},\mathbf{\Lambda}_{\mathrm{LR}},\mathbf{I}_{m}\right).
		\end{array}
	\end{equation*}
Besides, since both observations are acquired by different modality sensors, the noise, which is sensor-dependent, can be assumed statistically independent. Thus, $\mathbf{Y}_{\mathrm{HR}}^{t_i}|\mathbf{X}^{t_j},\Delta\mathbf{X}$ and $\mathbf{Y}_{\mathrm{LR}}^{t_j}|\mathbf{X}^{t_j}$ are also statistically independent and the joint likelihood function $p(\mathbf{Y}_{\mathrm{HR}}^{t_i},\mathbf{Y}_{\mathrm{LR}}^{t_j}|\mathbf{X}^{t_j},\Delta\mathbf{X})$ can be written as a simple product of the conditional distributions $p(\mathbf{Y}_{\mathrm{HR}}^{t_i}|\mathbf{X}^{t_j},\Delta\mathbf{X})$ and $p(\mathbf{Y}_{\mathrm{LR}}^{t_j}|\mathbf{X}^{t_j})$.
%
	
A Bayesian formulation of the robust multi-band image fusion problem allows prior information to be introduced to regularize the underlying estimation problem\cite{idierbayesian2008}. Bayesian estimators can be derived from the joint posterior distribution
	\begin{multline}
		 p(\mathbf{X}^{t_j},\Delta\mathbf{X}|\mathbf{Y}_{\mathrm{HR}}^{t_i},\mathbf{Y}_{\mathrm{LR}}^{t_j}) \propto \\ p(\mathbf{Y}_{\mathrm{HR}}^{t_i},\mathbf{Y}_{\mathrm{LR}}^{t_j}|\mathbf{X}^{t_j},\Delta\mathbf{X}) p(\mathbf{X}^{t_i}) p(\Delta\mathbf{X})
	\end{multline}
where $p(\mathbf{X}^{t_i})$ and $p(\Delta\mathbf{X})$ correspond to the prior distributions associated with the latent and change HR-HS images, respectively, assumed to be a priori independent. Under a maximum a posteriori (MAP) paradigm, the joint MAP estimator $\left\{\hat{\mathbf{X}}^{t_j}_{\mathrm{MAP}}, \mathbf{\Delta}\hat{\mathbf{X}}_{\mathrm{MAP}}\right\}$ can be derived by minimizing the negative log-posterior, leading to the following minimization problem
	\begin{equation}
		\label{eq:MAP}
		\left\{\hat{\mathbf{X}}^{t_i}_{\mathrm{MAP}}, \Delta\hat{\mathbf{X}}_{\mathrm{MAP}}\right\} \in \mathop{\rm Argmin}\limits_{\mathbf{X}^{t_j},\Delta\mathbf{X}}   \mathcal{J}\left(\mathbf{X}^{t_j},\Delta\mathbf{X}\right)
	\end{equation}
with
\begin{equation}
\label{eq:objective}
\begin{aligned}
  \mathcal{J}\left(\mathbf{X}^{t_j},\Delta\mathbf{X}\right)&=\frac{1}{2}\left\|\mathbf{\Lambda}_{\mathrm{HR}}^{-\frac{1}{2}} \left(\mathbf{Y}_{\mathrm{HR}}^{t_i} - \mathbf{L}\left(\mathbf{X}^{t_j}+\Delta\mathbf{X}\right)  \right) \right\|_{F}^{2}  \\
		&+ \frac{1}{2}\left\|\mathbf{\Lambda}_{\mathrm{LR}}^{-\frac{1}{2}} \left(\mathbf{Y}_{\mathrm{LR}}^{t_j} - \mathbf{X}^{t_j}\mathbf{BS} \right) \right\|_{F}^{2} \\
        &+ \lambda \phi_1\left(\mathbf{X}^{t_j}\right) + \gamma \phi_2 \left(\Delta\mathbf{X}\right).
\end{aligned}
\end{equation}
The regularizing functions $\phi_{1}(\cdot)$ and $\phi_{2}(\cdot)$ can be related to the negative log-prior distributions of the HR-HS latent and change images, respectively, and the parameters $\lambda$ and $\gamma$ tune the amount of corresponding penalizations in the overall objective function $\mathcal{J}(\mathbf{X}^{t_j},\Delta\mathbf{X})$. These functions should be carefully designed to exploit any prior knowledge regarding the parameters of interest. As discussed in Section \ref{subsec:robust_fusion}, numerous regularizations can be advocated for the HR-HS latent image $\mathbf{X}^{t_j}$. In this work, a Tikhonov regularization proposed in \cite{weibayesian2015-2} has been adopted
\begin{equation}
\label{eq:phi_1}
  \phi_1\left(\mathbf{X}^{t_j}\right) = \left\|\mathbf{X}^{t_j} - \bar{\mathbf{X}}^{t_j}\right\|_F^2
\end{equation}
where $\bar{\mathbf{X}}^{t_j}$ refers to a crude estimate of $\mathbf{X}^{t_j}$, e.g., resulting from a naive spatial interpolation of the observed LR-HS image $\mathbf{Y}_{\mathrm{LR}}^{t_j}$. This choice has been proven to maintain computational efficiency while providing accurate results \cite{loncanhyperspectral2015}. Additionally, a subspace-based representation can also be adopted to enforce $\mathbf{X}^{t_j}$ to live in a previously identified subspace, as advocated in \cite{weibayesian2015} and \cite{Simoes2014b}.

Conversely and more critically, a specific attention should be paid to the regularizing function $\phi_2(\cdot)$. This function should reflect the fact that most of the pixels are expected to remain unchanged in $\mathbf{X}^{t_i}$ and $\mathbf{X}^{t_j}$, i.e., most of the columns of the change image $\Delta \mathbf{X}$ are expected to be null vectors. This noticeable property can be easily translated by promoting the sparsity of the spectral change energy image  $\mathbf{e}$ defined by \eqref{eq:spectral_change_energy_image}. As a consequence, the regularizing function $\phi_2(\cdot)$ is chosen as the sparsity-inducing
$\ell_1$-norm of the change energy image $\mathbf{e}$ or, equivalently, as the $\ell_{2,1}$-norm of the change image
\begin{equation}
\label{eq:phi_2}
\phi_{2}\left(\Delta\mathbf{X}\right) = \left\|\Delta\mathbf{X}\right\|_{2,1} = \sum_{p=1}^{n} \left\|\Delta \mathbf{x}_p\right\|_2.
\end{equation}
This regularization is a specific instance of the non-overlapping group-lasso penalization \cite{Bach2011a} which has been considered in various applications to promote structured sparsity \cite{Cotter2005,Ding2006,Liu2009,wrightsparse2009,Nie2010,Lu2011,Fevotte2015}.

The next section describes an iterative algorithm which solves the minimization problem in \eqref{eq:MAP}.

\section{Minimization algorithm}
\label{sec:CD}

Computing the joint MAP estimator of the HR-HS latent image $\mathbf{X}^{t_j}$ at time $t_j$ and of the change image $\Delta\mathbf{X}$ can be achieved by solving the minimization problem in \eqref{eq:MAP}. However, no closed-form solution can be derived for this problem. Thus this section presents a minimization algorithm which iteratively converges to this solution. It consists in sequentially solving the problem w.r.t. to each individual variables $\mathbf{X}^{t_j}$ and $\Delta\mathbf{X}$. This block coordinate descent algorithm is summarized in Algo. \ref{BCD} whose main steps (fusion and correction) are detailed in what follows.

\begin{algorithm}[h!]
	\caption{BCD algorithm for robust multi-band image fusion}
	\label{alg:BCD}
	\begin{algorithmic}[1]
		\INPUT $\mathbf{Y}_{\mathrm{LR}}^{t_j}$, $\mathbf{Y}_{\mathrm{HR}}^{t_i}$, $\mathbf{L}$, $\mathbf{B}$, $\mathbf{S}$, $\mathbf{\Lambda}_{\mathrm{HR}}$, $\mathbf{\Lambda}_{\mathrm{LR}}$.
        \State Set $\Delta\mathbf{X}_{1}$.
		\For{$k = 1,\ldots,K$}
		\State $ \mathbf{X}^{t_j}_{k+1} = \argmin_{\mathbf{X}^{t_j}}\mathcal{J}(\mathbf{X}^{t_j},\Delta\mathbf{X}_{k})$
		\State $ \Delta\mathbf{X}_{k+1} =  \argmin_{\Delta\mathbf{X}}\mathcal{J}(\mathbf{X}^{t_j}_{k+1},\Delta\mathbf{X})$
		\EndFor
		\OUTPUT $\hat{\mathbf{X}}^{t_j}_{\mathrm{MAP}}\triangleq\mathbf{X}^{t_j}_{K+1}$ and $\Delta\hat{\mathbf{X}}_{\mathrm{MAP}}\triangleq \Delta\hat{\mathbf{X}}_{K+1}$
	\end{algorithmic}
	\label{BCD}
\end{algorithm}
			
	\subsection{Fusion: optimization w.r.t $\mathbf{X}^{t_j}$}
	
At the $k$th iteration of the BCD algorithm, let assume that the current value of the HR-HS change image is denoted $\Delta\mathbf{X}_{k}$. As suggested in Section \ref{subsec:robust_fusion}, an HR-PAN/MS corrected image ${\mathbf{Y}}_{\mathrm{cHR},k}^{t_j}$ that would be observed at time $t_j$ given the HR-PAN/MS image $\mathbf{Y}_{\mathrm{HR}}^{t_i}$ observed at time $t_i$ and the HR-HS change image $\Delta\mathbf{X}_{k}$ can be introduced as
	\begin{equation}
	\label{eq:pseudoObs1}
		{\mathbf{Y}}_{\mathrm{cHR},k}^{t_j} = \mathbf{Y}_{\mathrm{HR}}^{t_i} - \mathbf{L}\Delta\mathbf{X}_{k}.
	\end{equation}
Updating the current value of the HR-HS latent image consists in minimizing w.r.t. $\mathbf{X}^{t_j}$ the partial function
	\begin{equation}
	\label{eq:object1}
    \begin{aligned}
			\mathcal{J}_{1}\left(\mathbf{X}^{t_j}\right) &\triangleq \mathcal{J}\left(\mathbf{X}^{t_j},\Delta\mathbf{X}_{k}\right) \\
&=  \left\|\mathbf{\Lambda}_{\mathrm{LR}}^{-\frac{1}{2}} \left(\mathbf{Y}_{\mathrm{LR}}^{t_j} - \mathbf{X}^{t_j}\mathbf{BS}\right)\right\|_{F}^{2} \\
&+ \left\|\mathbf{\Lambda}_{\mathrm{HR}}^{-\frac{1}{2}} \left({\mathbf{Y}}_{\mathrm{cHR},k}^{t_j} - \mathbf{L}\mathbf{X}^{t_j}\right)\right\|_{F}^{2}
+ \lambda\phi_{1}\left(\mathbf{X}^{t_j}\right).
    \end{aligned}
	\end{equation}
As noticed earlier, this sub-problem boils down to the multi-band image fusion which has received considerable attention in the recent image processing and remote sensing literature \cite{Simoes2014b,loncanhyperspectral2015,weibayesian2015,weibayesian2015-2,weifast2015-2,weihyperspectral2015}. The two difficulties arising from this formulation lies in the high dimension of the optimization problem and in the fact that the sub-sampling operator $\mathbf{S}$ prevents any fast resolution in the frequency domain by diagonalization of the spatial degradation matrix $\mathbf{R}=\mathbf{B}\mathbf{S}$. However, with the particular choice \eqref{eq:phi_1} of the regularization function $\phi_1(\cdot)$ adopted in this paper, a closed-form solution can still be derived and efficiently implemented. It consists in solving a matrix Sylvester equation \cite{weifast2015-2} of the form
	\begin{equation}
		\mathbf{C}_{1}\mathbf{X}^{t_j} + \mathbf{X}^{t_j}\mathbf{C}_{2} = \mathbf{C}_{3}
	\end{equation}
where the matrices $\mathbf{C}_{1}$, $\mathbf{C}_{2}$ and $\mathbf{C}_{3}$ depend on the quantities involved in the problem, i.e., the virtual and observed images, the degradation operators, the noise covariance matrices and the spatially interpolated image defined in \eqref{eq:phi_1} (see \cite{weifast2015-2} for more details). Note that when a more complex regularization function $\phi_1(\cdot)$ is considered (e.g., TV or sparse representation over a dictionary), iterative algorithmic strategies can be adopted to approximate the minimizer of $\mathcal{J}_{1}\left(\mathbf{X}^{t_j}\right)$.

\subsection{Correction: optimization w.r.t $\Delta\mathbf{X}$}
	
Following the same strategy as in \cite{ferrarisdetecting2016}, let introduce the \emph{predicted} HR-PAN/MS image
\begin{equation}
\label{eq:pseudoObs2}
  {\mathbf{Y}}_{\mathrm{pHR},k}^{t_j} = \mathbf{L}\mathbf{X}^{t_j}_{k}
\end{equation}
that would be observed at time index $t_j$ by the HR-PAN/MS sensor given its spectral response $\mathbf{L}$ and the current state of the HR-HS latent image $\mathbf{X}^{t_j}_{k}$ at the $k$th iteration of the BCD algorithm. Similarly to \eqref{eq:assumption}, the predicted HR-PAN/MS change image can thus be defined as
	\begin{equation}
	\label{eq:pseudoObs3}
		\Delta{\mathbf{Y}}_{\mathrm{pHR},k} = \mathbf{Y}_{\mathrm{HR}}^{t_i} - {\mathbf{Y}}_{\mathrm{pHR},k}^{t_j}.
	\end{equation}
The objective function \eqref{eq:objective} w.r.t $\Delta\mathbf{X}$ is then rewritten by combining \eqref{eq:pseudoObs2} and \eqref{eq:pseudoObs3} with \eqref{eq:objective}, leading to
	\begin{equation}
	\label{eq:object2}
    \begin{aligned}
			\mathcal{J}_{2}(\Delta\mathbf{X})&\triangleq \mathcal{J}(\mathbf{X}^{t_j}_{k},\Delta\mathbf{X}) \\
&=  \left\|\mathbf{\Lambda}_{\mathrm{HR}}^{-\frac{1}{2}} \left(\Delta{\mathbf{Y}}_{\mathrm{pHR},k} - \mathbf{L}\Delta\mathbf{X}\right)\right\|_{F}^{2} + \gamma\phi_{2}\left(\Delta\mathbf{X}\right).
    \end{aligned}
		\end{equation}
With the specific CD-driven choice of $\phi_{2}\left(\cdot\right)$ in \eqref{eq:phi_2}, minimizing $\mathcal{J}_{2}(\Delta\mathbf{X})$ is an $\ell_{2,1}$-penalized least square problem. It is characterized by the sum of a convex and differentiable data fitting term with $\beta$-Lipschitz continuous gradient $\nabla\mathnormal{f}(\cdot)$
\begin{equation}
\label{eq:f_definition}
  f\left(\Delta\mathbf{X}\right) \triangleq \left\|\mathbf{\Lambda}_{\mathrm{HR}}^{-\frac{1}{2}} \left(\Delta{\mathbf{Y}}_{\mathrm{pHR},k} - \mathbf{L}\Delta\mathbf{X}\right)\right\|_{F}^{2}
\end{equation}
and a convex but non-smooth penalization
 \begin{equation}
    \label{eq:g_definition}
   g\left(\Delta\mathbf{X}\right) \triangleq \gamma \phi_2\left(\Delta\mathbf{X}\right) = \gamma \left\|\Delta\mathbf{X}\right\|_{2,1}.
 \end{equation}
Various algorithms have been proposed to solve such convex optimization problems including forward-backward splitting \cite{Combettes2005,Combettes2011}, Douglas-Rachford splitting \cite{Combettes2007,Combettes2011} and alternating direction method of multipliers \cite{Boyd2010,Parikh2013}. Since the proximal operator related to $g\left(\cdot\right)$ can be efficiently computed (see below), in this work, we propose to resort to an iterative forward-backward algorithm which has shown to provide the fastest yet reliable results. This algorithmic scheme is summarized in Algo. \ref{alg:FB}. It relies on a forward step which consists in conducting a gradient descent using the data-fitting function $f\left(\cdot\right)$ in \eqref{eq:f_definition}, and a backward step relying on the proximal mapping associated with the penalizing function $g\left(\cdot\right)$ in \eqref{eq:g_definition}.


\begin{algorithm}[h!]
		\caption{Correction step: forward-backward algorithm}
		\label{alg:FB}
		\begin{algorithmic}[2]
        \INPUT $\Delta\mathbf{X}_{k}$, $\Delta{\mathbf{Y}}_{\mathrm{pHR},k}$, $\mathbf{\Lambda}_{\mathrm{HR}}$, $\mathbf{L}$, $\left\{\eta_j\right\}_{j=1}^{J}$
        \State Set $\mathbf{V}_{1}\triangleq \Delta\mathbf{X}_{k}$
		\For{$j=1,\ldots,J$}
        \State \emph{\% forward step}
        \State $\mathbf{U}_{j+1} = \mathbf{V}_{j}-\eta_{j} \nabla f\left(\mathbf{V}_{j}\right)$ \label{algo:step_forward}
        \State \emph{\% backward step}
		\State $\mathbf{V}_{j+1} = \mathrm{prox}_{\eta_j g}\left(\mathbf{U}_{j+1}\right)$ \label{algo:step_backward}
		\EndFor
		\OUTPUT $\Delta\mathbf{X}_{k+1}\triangleq \mathbf{V}_{J+1}$
		\end{algorithmic}
\end{algorithm}

Since the HR-PAN/MS observed image has only a few spectral bands (e.g., $n_{\lambda} \sim 10$), the spectral degradation matrix $\mathbf{L}\in \mathbb{R}^{n_{\lambda} \times m_{\lambda}}$ is a fat (and generally full-row rank) matrix. Thus, the corresponding gradient operator $\nabla f\left(\cdot\right)$ defining the forward step (see line \ref{algo:step_forward} of Algo. \ref{alg:FB}) can be easily and efficiently computed. Conversely, the proximal operator associated with $g(\cdot)$ in \eqref{eq:g_definition} and required during the backward step (see line \ref{algo:step_backward} of Algo. \ref{alg:FB}) is defined as	
		\begin{equation}
\label{eq:proxU}
		\textrm{prox}_{\eta \mathnormal{g}}(\mathbf{U}) = \underset{\mathbf{Z}}{\argmin}\left(\gamma\left\|\mathbf{Z}\right\|_{2,1} + \frac{1}{2\eta}\left\|\mathbf{Z}-\mathbf{U}\right\|_{F}^{2}\right)
		\end{equation}
for some $\eta>0$. The function $\mathnormal{g}(\mathbf{U})$ in \eqref{eq:g_definition} can be split as $\sum_{p=1}^{n} \mathnormal{g}_{p}(\mathbf{u}_p)$ with, for each column, $\mathnormal{g}_{p}(\cdot) = \gamma \left\|\cdot\right\|_{2}$. Based on the separability property of proximal operators \cite{Parikh2013}, the operator \eqref{eq:proxU} can be decomposed and computed for each pixel location $p$ ($p=1,\ldots,n$) as	
		\begin{equation}
        \label{eq:prox_split}
		\left[\textrm{prox}_{\eta g}(\mathbf{U})\right]_{p} = \textrm{prox}_{\eta g_p}(\mathbf{u}_{p})
		\end{equation}
where the notations $\left[\cdot\right]_p$ stands for the $p$th column. Thus, only the proximal operator associated with the Euclidean distance induced by the  $\ell_2$-norm is necessary. The Moreau decomposition \cite{Parikh2013}
		\begin{equation}
		\label{eq:Moreau}
		\mathbf{u}_p = \text{prox}_{\eta\mathnormal{g}}\left(\mathbf{u}_p\right) + \eta\text{prox}_{\eta^{-1} g_p^*}\left(\eta^{-1}\mathbf{u}_p\right)
		\end{equation}
establishes a relationship between the proximal operators of the function $g_p(\cdot)$ and its conjugate $g_p^*(\cdot)$. When the function $g(\cdot)$ is a general norm, its conjugate corresponds to the indicator function into the ball $\mathbb{B}$ defined by its dual norm \cite{wrightsparse2009,Parikh2013}, leading to
	\begin{equation}
		\label{eq:MoreauProj}
		\text{prox}_{\eta\mathnormal{g}}(\mathbf{u}_p) = \mathbf{u}_p - \eta \mathcal{P}_{\mathbb{B}}\left(\frac{\mathbf{u}_p}{\eta}\right)
		\end{equation}
where $\mathcal{P}_{\mathbb{B}}(\cdot)$ denotes the projection. When $g(\cdot)$ is defined by \eqref{eq:g_definition}, since the  $\ell_2$-norm is self-dual, this projection is
	\begin{equation}
		\label{eq:euclidianBall}
			\mathcal{P}_{\mathbb{B}}\left(\mathbf{u}_p\right) = \left\{\begin{array}{ll}			
			\frac{\gamma\mathbf{u}_p}{\left\|\mathbf{u}_p\right\|_{2}} & \text{if} \left\|\mathbf{u}_p\right\|_{2}> \gamma\\
			\mathbf{u}_p & \text{otherwise.}
			\end{array}
			\right.
	\end{equation}
Consequently, replacing \eqref{eq:euclidianBall} in \eqref{eq:MoreauProj}, the proximal operator associated with the function $g_p(\cdot)$ in \eqref{eq:prox_split} is
		\begin{equation}
		\label{eq:proxReg}
		\text{prox}_{\eta\mathnormal{g_p}}(\mathbf{u}_p) = \left\{\begin{array}{ll}			
				\left(1 - \frac{\eta\gamma}{\left\|\mathbf{u}_p\right\|_{2}}\right)\mathbf{u}_p & \text{if} \left\|\mathbf{u}_p\right\|_{2}>\eta\gamma\\
				0 & \text{otherwise.}
				\end{array}
			\right.
	\end{equation}
To conclude, the correction procedure can be interpreted as first a gradient descent step for spectral deblurring of the HR-HS change image from the HR-PAN/MS predicted change image (forward step), followed by a soft-thresholding of the resulting HR-HS change image to promote spatial sparsity (backward step).

\newcommand{\subfigwidth}{0.19\textwidth}

	\begin{figure*}%
			\begin{subfigure}[b]{\subfigwidth}
					\centering	
					\includegraphics[trim={0 0 0 0},clip,scale=0.255]{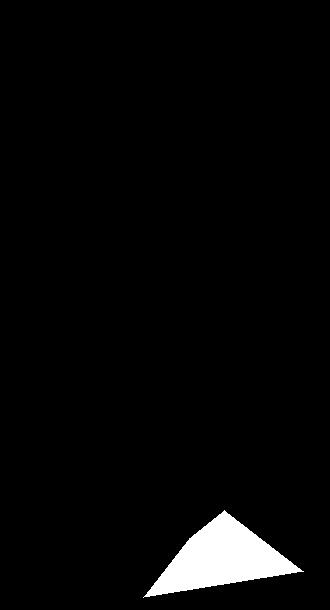}
					\caption{$\mathbf{d}_{\mathrm{HR}}$}
					\label{fig:chMap_true}
			\end{subfigure}
			\begin{subfigure}[b]{\subfigwidth}
					\centering	
					\includegraphics[trim={0 0 0 0},clip,scale=0.255]{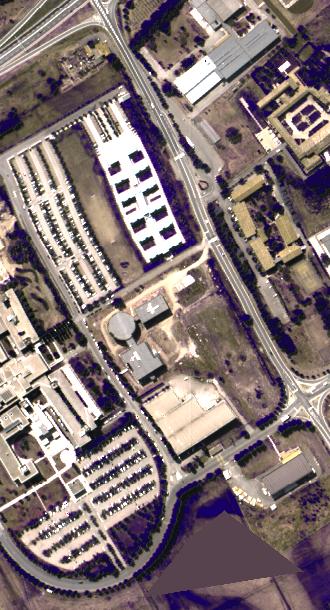}
					\caption{$\mathbf{X}^{t_i}$}
					\label{fig:ch1}
			\end{subfigure}
			\begin{subfigure}[b]{\subfigwidth}
					\centering	
					\includegraphics[trim={0 0 0 0},clip,scale=0.255]{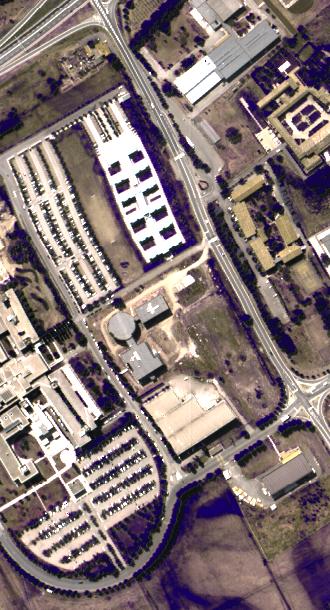}
					\caption{$\mathbf{X}^{t_j}$}
					\label{fig:ref1}
			\end{subfigure}
			\begin{subfigure}[b]{\subfigwidth}
					\centering	
					\includegraphics[trim={0 0 0 0},clip,scale=0.255]{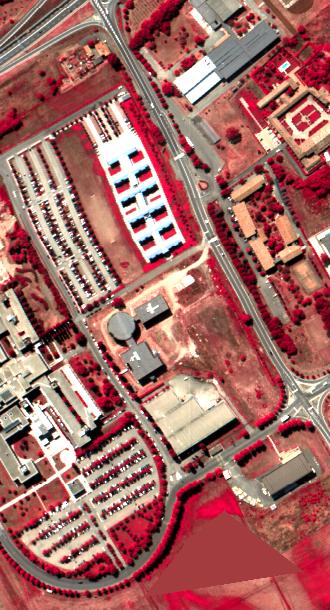}
					\caption{$\mathbf{Y}_{\mathrm{HR}}^{t_i}$}
					\label{fig:ms1}
			\end{subfigure}
			\begin{subfigure}[b]{\subfigwidth}
					\centering	
					\includegraphics[trim={1.05cm 0.05cm 1cm 0.15cm},clip,scale=1.27]{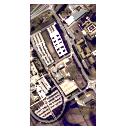}
					\caption{$\mathbf{Y}_{\mathrm{LR}}^{t_j}$}
					\label{fig:hs1}
			\end{subfigure}
			\caption{One particular simulation configuration: \subref{fig:chMap_true} the HR change mask $\mathbf{d}_{\mathrm{HR}}$ , \subref{fig:ch1}-\subref{fig:ref1} the HR-HS latent images $\mathbf{X}^{t_i}$ and $\mathbf{X}^{t_j}$, {\subref{fig:ms1}}-\subref{fig:hs1} the spectrally degraded version HR-MS observed image $\mathbf{Y}_{\mathrm{HR}}^{t_i}$  and the spatially degraded LR-HS observed image $\mathbf{Y}_{\mathrm{LR}}^{t_j}$. Note that, in this particular configuration, the change-inducing function $\vartheta^{t_j}(\cdot,\mathbf{d}_{\mathrm{HR}})$ is the identity operator (i.e., $\mathbf{A}^{t_j}=\mathbf{A}^{\mathrm{ref}}$) since it does not apply any change into the corresponding HR-HS latent image $\mathbf{X}^{t_j}$ while the function $\vartheta^{t_i}(\cdot,\mathbf{d}_{\mathrm{HR}})$ includes a triangular region of pixels in $\mathbf{X}^{t_i}$ affected by changes. Moreover, the HR observed image is here an MS image.  \label{fig:CONF}}
\end{figure*}

\section{Experiments}
\label{sec:experiments}

\subsection {Simulation framework}
	
Real dataset for assessing performance of CD algorithms is rarely available. Indeed, this assessment requires a couple of images acquired at two different dates, geometrically and radiometrically pre-corrected, presenting changes and, for the scenario considered in this paper, coming from two different optical sensors. To alleviate this issue, inspired by the well-known Wald's evaluation protocol dedicated to pansharpening algorithms \cite{waldfusion1997}, a framework has been proposed in \cite{ferrarisdetecting2016} to assess the performance of CD algorithms when dealing with optical images of different spatial and spectral resolutions. This framework only requires a single HR-HS reference image $\mathbf{X}^{\mathrm{ref}}$ and generates a pair of latent HR-HS images $\mathbf{X}^{t_i}$ and $\mathbf{X}^{t_j}$ resulting from a unmixing-mixing process. This process allows synthetic yet realistic changes to be incorporated within one of these latent images, w.r.t. a pre-defined binary reference HR change mask $\mathbf{d}_{\mathrm{HR}} \in \mathbb{R}^n$ locating the pixels affected by these changes and further used to assess the performance of the CD algorithms. This procedure allows various physically-inspired changes to be considered, e.g., by tuning the relative abundance of a each endmember or replacing one of them my another. This protocol is briefly described below (see \cite{ferrarisdetecting2016} for more details).

\subsubsection{Reference image}

The HR-HS reference image $\mathbf{X}^{\mathrm{ref}}$  used in the experiments reported in this paper is a $610 \times 330 \times 115$ HS image of the Pavia University, Italy, acquired by the reflective optics system imaging spectrometer (ROSIS) sensor. This image has undergone a pre-precessing to smooth the atmospheric effects of vapor water absorption by removing some bands. Thus the final HR-HS reference image is of size $610 \times 330 \times 93$.
	
\subsubsection{Generating the changes}

Using the same procedure proposed in \cite{ferrarisdetecting2016}, the HR-HS reference image $\mathbf{X}^{\mathrm{ref}} \in \mathbb{R}^{m_{\lambda}\times n}$ has been linearly unmixed to define the reference matrix $\mathbf{M}^{\mathrm{ref}} \in \mathbb{R}^{m_{\lambda}\times R}$ of $R$ endmember spectral signatures and the corresponding reference abundance matrix $\mathbf{A}^{\mathrm{ref}} \in \mathbb{R}^{R\times n}$ such that $\mathbf{X}^{\mathrm{ref}}\approx \mathbf{M}^{\mathrm{ref}} \mathbf{A}^{\mathrm{ref}}$. The two latent HR-HS images $\mathbf{X}^{t_i}$ and $\mathbf{X}^{t_j}$ are then computed as linear mixture of the endmembers in $\mathbf{M}^{\mathrm{ref}}$ with corresponding abundance matrices $\mathbf{A}^{t_i}$ and $\mathbf{A}^{t_j}$, respectively, derived from the reference abundances $\mathbf{A}^{\mathrm{ref}}$ and the change mask $\mathbf{d}_{\mathrm{HR}}$, i.e.,
\begin{align}
  \mathbf{X}^{t_i}= \mathbf{M}^{\mathrm{ref}}\mathbf{A}^{t_i} & \quad \text{and} \quad \mathbf{X}^{t_j}= \mathbf{M}^{\mathrm{ref}}\mathbf{A}^{t_j}\nonumber
\end{align}
with
\begin{align}
\mathbf{A}^{t_i} = \vartheta^{t_i}\left(\mathbf{A}^{\mathrm{ref}},\mathbf{d}_{\mathrm{HR}}\right) & \quad \text{and} \quad \mathbf{A}^{t_j} = \vartheta^{t_j}\left(\mathbf{A}^{\mathrm{ref}},\mathbf{d}_{\mathrm{HR}}\right) \nonumber
\end{align}
where the two change-inducing functions $\vartheta^{t_{\cdot}}: \mathbb{R}^{R\times n}\times \mathbb{R}^n \rightarrow \mathbb{R}^{R\times n}$ are defined to simulate realistic changes in some pixels of the HR-HS latent images. Three sets of $75$ predefined change masks have been designed according to three specific change rules introduced in \cite{ferrarisdetecting2016}. For each simulated pair $\left\{\mathbf{X}^{t_i},\mathbf{X}^{t_j}\right\}$, one of the two functions $\vartheta^{t_\cdot}(\cdot,\mathbf{d}_{\mathrm{HR}})$ is defined as a ``no-change'' operator, i.e., $\vartheta^{t_\cdot}(\mathbf{A}^{\mathrm{ref}},\mathbf{d}_{\mathrm{HR}}) = \mathbf{A}^{\mathrm{ref}}$, which leads to an overall set of $450$ simulated pairs $\left\{\mathbf{X}^{t_i},\mathbf{X}^{t_j}\right\}$ of HR-HS latent images.

\subsubsection{Generating the observed images}

The HR-PAN/MS observed image $\mathbf{Y}_{\mathrm{HR}}^{t_i}$ is obtained by spectrally degrading the corresponding HR-HS latent image $\mathbf{X}^{t_i}$. Two scenarios are considered. Scenario 1 consists in averaging the first $43$ bands of the HR-HS latent image to produce an HR-PAN image. Conversely, Scenario 2 considers an HR-MS image by spectrally filtering the HR-HS latent image $\mathbf{X}^{t_i}$ with a $4$-band LANDSAT-like spectral response. Moreover, to generate a spatially degraded image $\mathbf{Y}_{\mathrm{LR}}^{t_j}$, the respective latent image $\mathbf{X}^{t_i}$ has been blurred by a $5 \times 5$ Gaussian kernel and subsequently equally down-sampled in the vertical and horizontal directions with a down-sampling ratio $d = 5$. To illustrate, Fig. \ref{fig:CONF} shows one of the $450$ simulation configurations used during the experiments.

\subsection{Compared methods}	
	
The proposed robust fusion-based CD technique has been compared to four methods able to deal with optical images of different spatial and spectral resolutions. The first one has been proposed in \cite{ferrarisdetecting2016} and also relies on a fusion-based approach. Up to the authors' knowledge, it was the first operational CD technique able to operate with multi-band optical images of different spatial and spectral images. Contrary to the model \eqref{eq:jointobsmodel} proposed in this paper, it consists in recovering a common latent image by fusing the two observed images and then predicting an HR-PAN/MS image $\hat{\mathbf{Y}}_{\mathrm{HR}}^{\mathrm{F},t_i}$ from the underlying forward model. An HR-PAN/MS change image $\Delta {\mathbf{Y}}_{\mathrm{HR}}^{\mathrm{F},t_i}  $ has been then computed as in \eqref{eq:assumption} from the pair of HR-PAN/MS observed and predicted images $\left\{{\mathbf{Y}}_{\mathrm{HR}}^{t_i} , \hat{\mathbf{Y}}_{\mathrm{HR}}^{\mathrm{F},t_i}\right\}$. Finally, as recommended in \cite{ferrarisdetecting2016}, a spatially-regularized CVA (sCVA) similar to the decision rule detailed in Section \ref{subsec:ps} has been conducted on $\Delta {\mathbf{Y}}_{\mathrm{HR}}^{\mathrm{F},t_i}$ to produce an estimated HR CD mask denoted $\hat{\mathbf{d}}_{\mathrm{F}}$.

The second method aims at producing an HR-PAN/MS predicted image by successive spatial superresolution and spectral degradation. More precisely, an HR-HS latent image is first recovered by conducting a band-wise spatial superresolution of the observed LR-HS ${\mathbf{Y}}_{\mathrm{LR}}^{t_j}$ following the fast method in \cite{zhaofast2016}. Then this latent image is spectrally degraded according to produce an HR-PAN/MS predicted image $\hat{\mathbf{Y}}_{\mathrm{HR}}^{\mathrm{SD},t_j}$. Similarly to the previous fusion-based method, sCVA has been finally conducted on the pair $\left\{{\mathbf{Y}}_{\mathrm{HR}}^{t_i}, \hat{\mathbf{Y}}_{\mathrm{HR}}^{\mathrm{SD},t_j}\right\}$ to produce an HR CD mask denoted $\hat{\mathbf{d}}_{\mathrm{SD}}$. The third CD method applies the same procedure with a reverse order of spatial superresolution and spectral degradation, and produces produces an HR change mask denoted $\hat{\mathbf{d}}_{\mathrm{DS}}$ from the pair of HR-PAN/MS images $\left\{{\mathbf{Y}}_{\mathrm{HR}}^{t_i} , \hat{\mathbf{Y}}_{\mathrm{HR}}^{\mathrm{DS},t_j}\right\}$. The fourth CD method, referred to as the worst-case (WC) as in \cite{ferrarisdetecting2016}, build a LR change mask $\hat{\mathbf{d}}_{\mathrm{WC}}$ by crudely conducting a sCVA on a spatially degraded version of the HR-PAN/MS image and a spectrally degraded version of the LR-HS image.

\subsection{Figures-of-merit}	
\label{subsec:figures_of_merit}
The CD performances of these four methods, as well as the performance of the proposed robust fusion-based method whose HR change mask is denoted $\hat{\mathbf{d}}_{\mathrm{RF}}$, have been visually assessed from empirical receiver operating characteristics (ROC), representing the estimated pixel-wise probability of detection ($\mathrm{PD}$) as a function of the probability of false alarm ($\mathrm{PFA}$). Moreover, two quantitative criteria derived from these ROC curves have been computed, namely, $i$) the area under the curve (AUC), corresponding to the integral of the ROC curve and $ii$) the distance between the no detection point $(PFA = 1, \mathrm{PD} = 0)$ and the point at the interception of the ROC curve with the diagonal line defined by $\mathrm{PFA} = 1 - \mathrm{PD}$. For both metrics, greater the criterion, better the detection.

%

\begin{figure}[h]%
	\centering	
    \includegraphics[trim={0.6cm 3.7cm 1.6cm 3cm},clip,width=0.9\columnwidth]{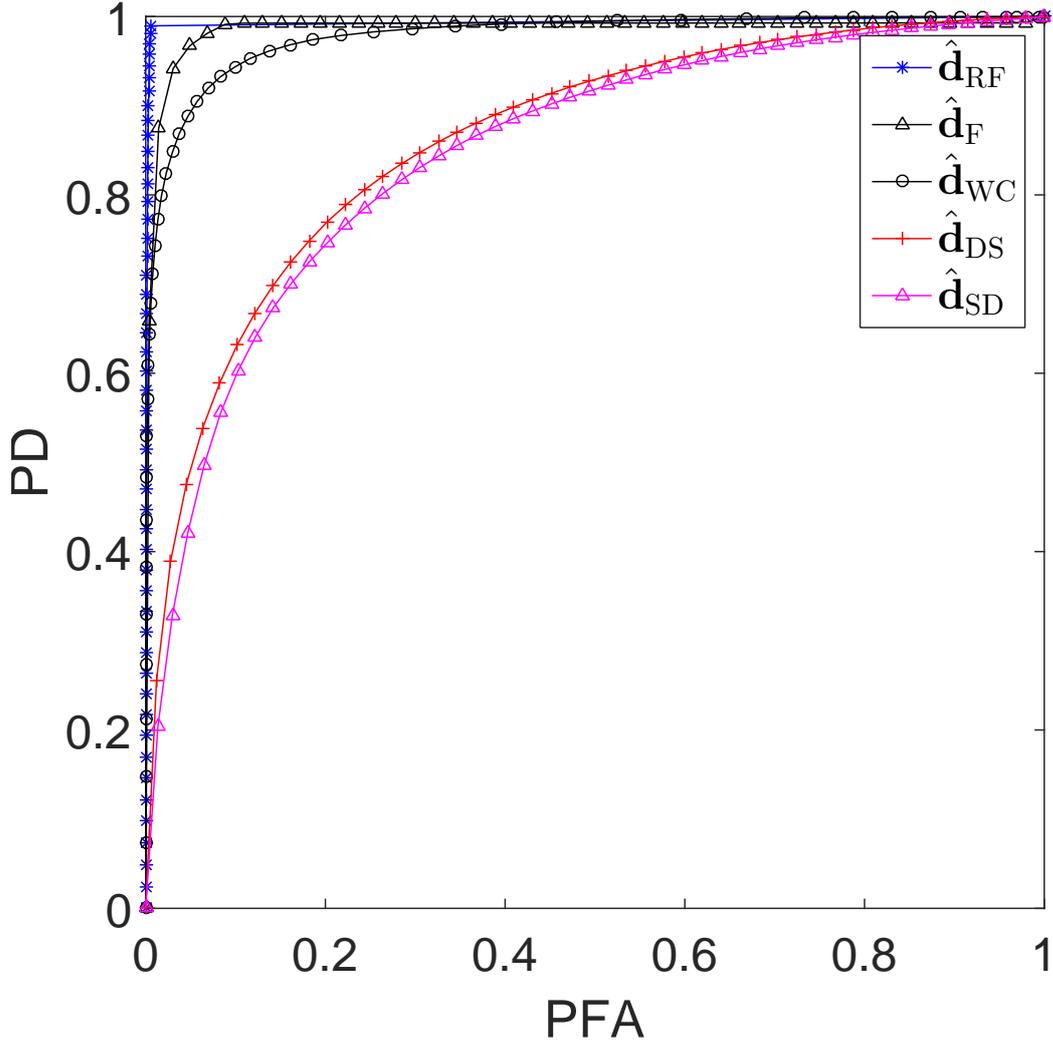}\vspace{-0.8cm}
	\caption{Scenario 1 (HR-PAN vs. LR-HS): ROC curves.}%
	 \label{fig:rocPANHS}%
\end{figure}

\newcommand{\one}[1]{\bf{\textcolor[rgb]{0.00,0.00,1.00}{#1}}}
\newcommand{\two}[1]{\textcolor[rgb]{0.00,0.00,1.00}{#1}}
\setlength{\tabcolsep}{5pt}
\renewcommand{\arraystretch}{1.3}

\begin{table}[h!]
    \caption{Scenarios 1 \& 2: quantitative detection performance (AUC and distance).}
    \centering
    \begin{tabular}{|c|c|c|c|c|c|c|c|}
    \cline{3-7}
    \multicolumn{2}{c|}{} & $\hat{\mathbf{d}}_{\mathrm{RF}}$ & $\hat{\mathbf{d}}_{\mathrm{RF}}$ & $\hat{\mathbf{d}}_{\mathrm{WC}}$ & $\hat{\mathbf{d}}_{\mathrm{DS}}$ & $\hat{\mathbf{d}}_{\mathrm{SD}}$\\
    \hline
		\hline
		\multirow{2}{*}{\rotatebox{00}{Scenario 1}}      & AUC   & $\one{0.9936}$ & $\two{0.9853}$& $0.9777$& $0.8623$& $0.8469$\\
                                                     & Dist. & $\one{0.9896}$ & $\two{0.9578}$& $0.9249$& $0.7832$& $0.7716$\\
    \hline
    \multirow{2}{*}{\rotatebox{00}{Scenario 2}}      & AUC   & $\one{0.9974}$ & $\two{0.9919}$& $0.9809$& $0.8881$& $0.8915$\\
                                                     & Dist. & $\one{0.9944}$ & $\two{0.9590}$& $0.9356$& $0.8141$& $0.8197$\\
    \hline
    \end{tabular}
  \label{table:ROCSEN}
\end{table}

\subsection{Results}

\subsubsection{Scenario 1 (HR-PAN vs. LR-HS)}
	
The ROC curves depicted in Fig. \ref{fig:rocPANHS} with corresponding metrics in Table \ref{table:ROCSEN} (first two rows) correspond to the CD results obtained from a pair of HR-PAN and LR-HS observed images. Clearly, the proposed robust fusion-based CD technique outperforms the four other CD techniques. More importantly, it provides almost perfect detections even for very low PFA, i.e., for very low energy changes. Note that the CD mask $\mathbf{d}_{\mathrm{WC}}$ estimated by the worst-case method is defined at an LR.

%
	
\begin{figure}[h!]%
	\centering	
	\includegraphics[trim={0.6cm 3.7cm 1.6cm 3cm},clip,width=0.9\columnwidth]{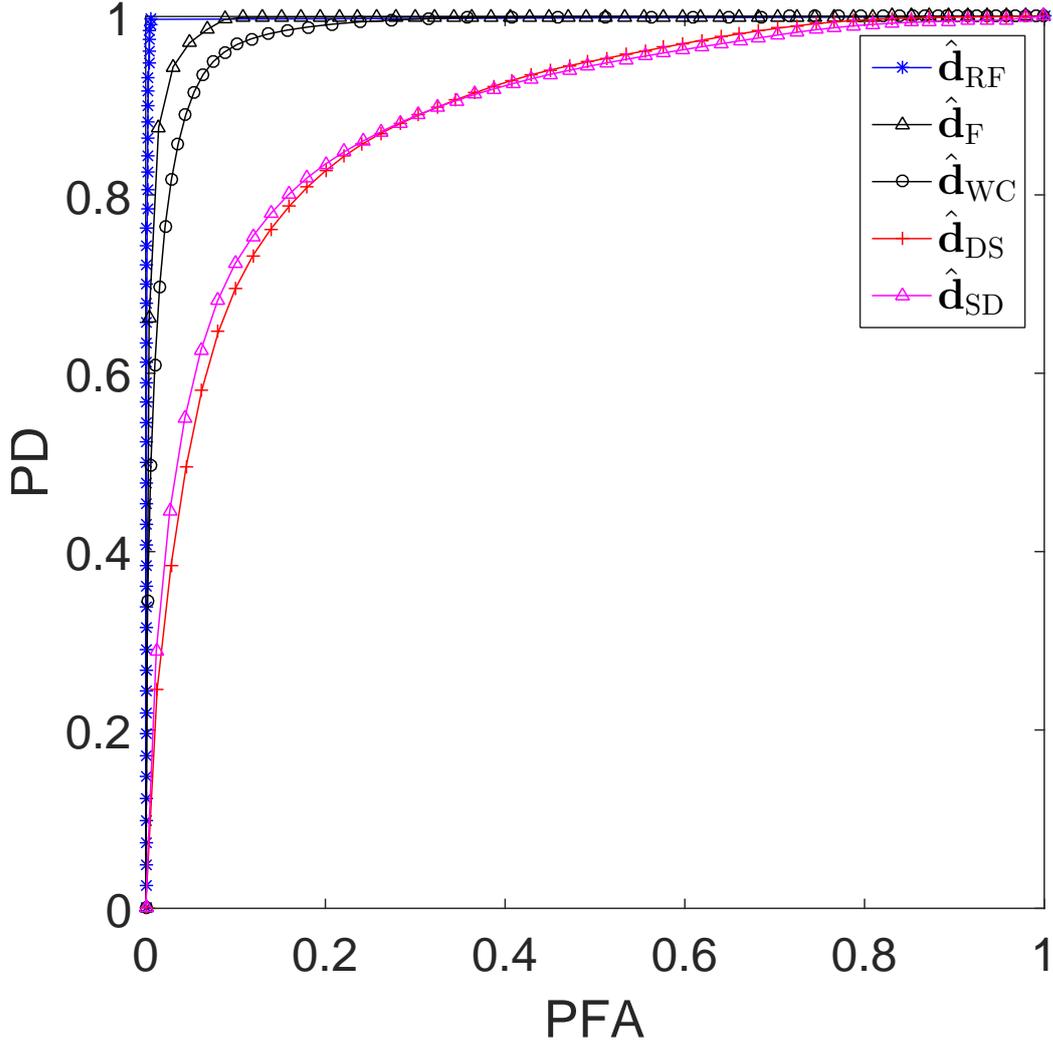}\vspace{-0.8cm}
			\caption{Scenario 2 (HR-MS vs. LR-HS): ROC curves. }
	 \label{fig:rocMSHS}%
\end{figure}

\subsubsection{Scenario 2 (HR-MS vs. LR-HS)}

Applying the same procedure of Scenario 1 but now considering an HR-MS observed image instead of the HR-PAN observed image leads to very similar overall performance. The ROC plot is depicted in Fig. \ref{fig:rocMSHS} with corresponding metrics in Table \ref{table:ROCSEN} (last two rows). As in Scenario 1, comparing curves in Fig. \ref{fig:rocPANHS} shows that the proposed method offers a higher precision even when analyzing a lower spectral resolution HR observed image.

\begin{figure}[h!]
			\begin{subfigure}[b]{0.24\textwidth}
					\centering	
					\includegraphics[trim={2.3cm 2cm 2.3cm 0.7cm},clip,scale=0.242]{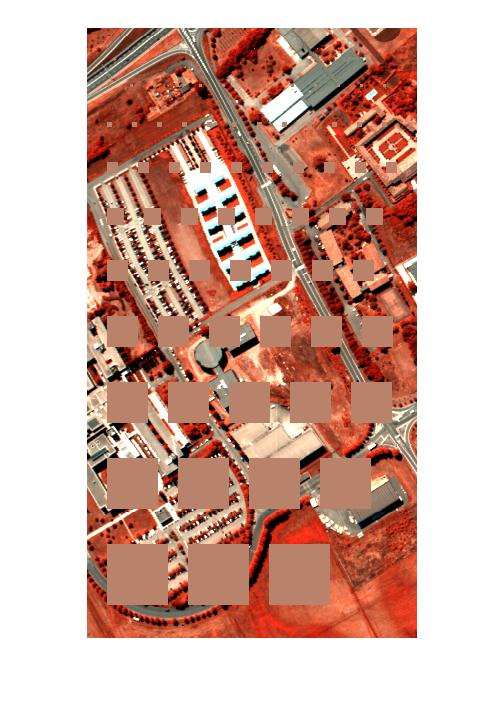}
					\caption{$\mathbf{Y}_{\mathrm{HR}}^{t_i}$}
					\label{fig:obsMS}
			\end{subfigure}
			\begin{subfigure}[b]{0.24\textwidth}
					\centering	
					\includegraphics[trim={3.05cm 2.3cm 3.05cm 1cm},clip,scale=1.21]{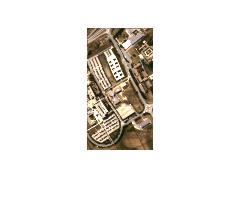}
					\caption{$\mathbf{Y}_{\mathrm{LR}}^{t_j}$}
					\label{fig:obsHS}
			\end{subfigure}
			\begin{subfigure}[b]{0.24\textwidth}
					\centering	
					\includegraphics[trim={2.3cm 1.42cm 2.3cm 0.7cm},clip,scale=0.32]{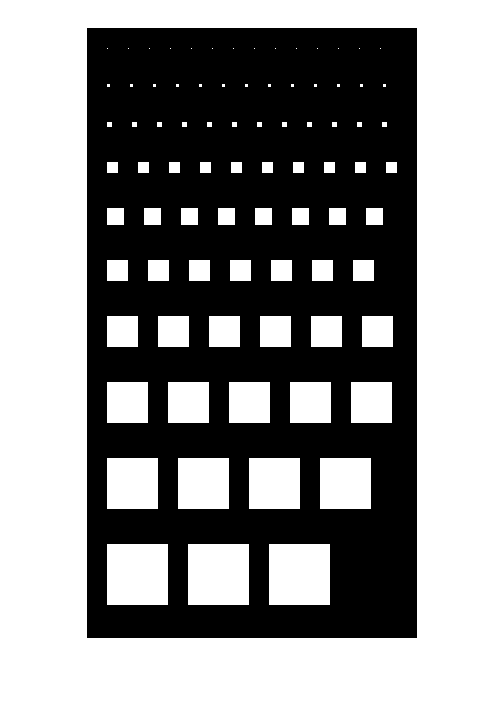}
					\caption{$\mathbf{d}_{\mathrm{HR}}$}
					\label{fig:maskHighRes}
			\end{subfigure}
			\begin{subfigure}[b]{0.24\textwidth}
					\centering	
					\includegraphics[trim={2.3cm 1.42cm 2.3cm 0.7cm},clip,scale=0.32]{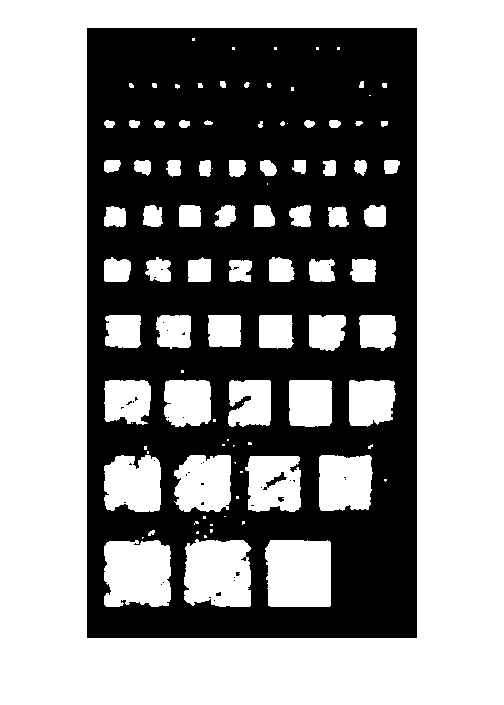}
					\caption{$\hat{\mathbf{d}}_{\mathrm{F}}$}
					\label{fig:changeDetectionFus}
			\end{subfigure}
			\begin{subfigure}[b]{0.24\textwidth}
					\centering	
					\includegraphics[trim={2.3cm 1.7cm 2.3cm 0.7cm},clip,scale=1.6]{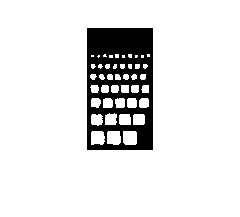}
					\caption{$\hat{\mathbf{d}}_{\mathrm{WC}}$}
					\label{fig:changeDetectionWorst}
			\end{subfigure}
			\begin{subfigure}[b]{0.24\textwidth}
					\centering	
					\includegraphics[trim={2.3cm 2cm 2.3cm 0.7cm},clip,scale=0.242]{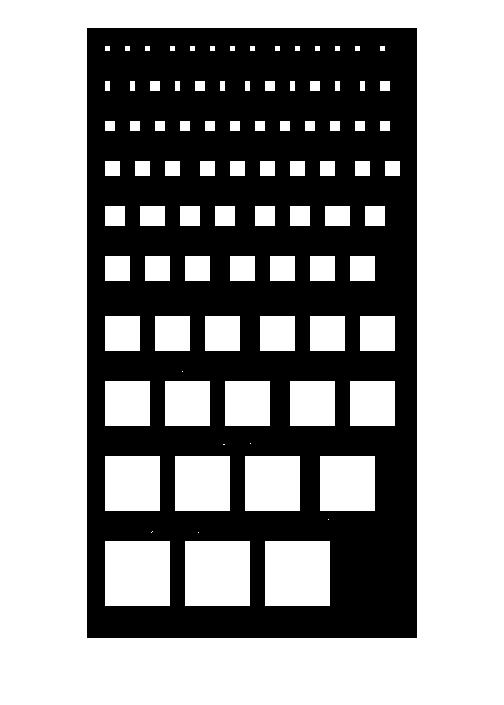}
					\caption{$\hat{\mathbf{d}}_{\mathrm{RF}}$}
					\label{fig:changeDetectionProp}
			\end{subfigure}
			\caption{CD precision for Scenario 2 (HR-MS vs. LR-HS): \protect\subref{fig:obsMS} HR-MS observed image $\mathbf{Y}_{\mathrm{HR}}^{t_i}$, \protect\subref{fig:obsHS} LR-HS observed image $\mathbf{Y}_{\mathrm{LR}}^{t_j}$, \protect\subref{fig:maskHighRes} actual change mask $\mathbf{d}_{\mathrm{HR}}$, \protect\subref{fig:changeDetectionFus} change mask $\hat{\mathbf{d}}_{\mathrm{F}}$ estimated by the fusion-based approach \cite{ferrarisdetecting2016}, \protect\subref{fig:changeDetectionWorst} change mask $\hat{\mathbf{d}}_{\mathrm{WC}}$ estimated by the worst-case approach, \protect\subref{fig:changeDetectionProp} change mask $\hat{\mathbf{d}}_{\mathrm{RF}}$ estimated by the proposed robust fusion-based approach.}%
	\label{fig:comp}%
\end{figure}

As an additional result, for Scenario 2, Fig. \ref{fig:comp} compares the abilities of detecting changes of decreasing size of the proposed method against the fusion-based CD method  \cite{ferrarisdetecting2016} and the worst-case CD method. Figure \ref{fig:obsMS} and \ref{fig:obsHS} shows the observed image pair $\mathbf{Y}_{\mathrm{HR}}^{t_i}$ and $\mathbf{Y}_{\mathrm{LR}}^{t_j}$ containing multiple changes with size varying from $1\times 1$- pixel to $61 \times 61$-pixels, with the corresponding change mask $\mathbf{d}_{\mathrm{HR}}$ presented in Fig. \ref{fig:maskHighRes}. Figures \ref{fig:changeDetectionFus} and \ref{fig:changeDetectionWorst} present the change masks $\hat{\mathbf{d}}_{\mathrm{F}}$ and $\hat{\mathbf{d}}_{\mathrm{WC}}$ recovered by the two competing methods, respectively, while the CD mask $\hat{\mathbf{d}}_{\mathrm{RF}}$ recovered by the proposed robust fusion-based method is reported in Fig. \ref{fig:changeDetectionProp} shows the proposed CD. For each technique, the decision threshold $\tau$ required in the CVA in \eqref{eq:CVArule} has been tuned to reach the higher distance value in the corresponding ROC curves. The first advantage of the proposed method is a significant decrease of the number of false alarm which are due to propagated errors when implementing the two other methods. Moreover, these results proves once again that the proposed method achieves a better detection rate with a higher resolution, even when considering extremely localized change regions. Remaining false alarms only occur near edges between change and no-change regions of small size due to the difference of spatial resolutions and the width of the blur kernel. Note also that the CD mask estimated by the worst-case method is of coarse scale since based on the comparison of two LR-MS images.

\section{Conclusion}

This paper proposed a robust fusion-based change detection technique to handle two multi-band optical observed images of different spatial and spectral resolutions. The technique was based on the definition of two high resolution hyperspectral latent images related to the observed images via a double physically-inspired forward model. The difference between these two latent images was assumed to be spatially sparse, implicitly locating the changes at a high resolution scale. Inferring these two latent images was formulated as an inverse problem which was solved within a $2$-step iterative scheme. This algorithmic strategy amounted to solve a standard fusion problem and an $\ell_{2,1}$-penalized spectral deblurring step. Contrary to the methods already proposed in the literature, modeling errors were not anymore propagate in-between steps. A simulation protocol allowed the performance of the proposed technique in terms of detection and precision to be assessed and compared with the performance of various algorithms. The detection rate as well as the accuracy of this method was clearly improved by this robust-fusion based algorithm. Future works include the detection of change between optical and non-optical data.

\bibliographystyle{ieeetran}
\bibliography{strings_all_ref,HSbib_cleaned}

\end{document}